\documentclass[letterpaper]{article} 
\usepackage{aaai2026}  
\usepackage{times}  
\usepackage{helvet}  
\usepackage{courier}  
\usepackage[hyphens]{url}  
\usepackage{graphicx} 
\usepackage{natbib}  
\usepackage{caption} 
\usepackage{algorithm}
\usepackage{algorithmic}

\usepackage{newfloat}
\usepackage{listings}
\DeclareCaptionStyle{ruled}{labelfont=normalfont,labelsep=colon,strut=off} 
\floatstyle{ruled}
\newfloat{listing}{tb}{lst}{}
\floatname{listing}{Listing}

\usepackage{amsmath}
\usepackage{multirow}
\usepackage{booktabs} 
\usepackage{colortbl}
\usepackage{subfigure}
\usepackage[table]{xcolor}
\definecolor{light-gray}{gray}{0.7}
\definecolor{light-gray0}{gray}{0.92}
\usepackage{amsthm}

\usepackage{amssymb}
\usepackage{tcolorbox}

\title{User-centric Subjective Leaderboard by Customizable Reward Modeling}
\author{
    Qi Jia\textsuperscript{\rm 1}, 
    Xiujie Song\textsuperscript{\rm 2}, 
    Zicheng Zhang\textsuperscript{\rm 1,2}, 
    Yijin Guo\textsuperscript{\rm 1,2},\\ 
    Kaiwei Zhang\textsuperscript{\rm 2}, 
    Zijian Chen\textsuperscript{\rm 2}, 
    Guangtao Zhai\textsuperscript{\rm 1,2}\thanks{Corresponding author}
}
\affiliations{
    \textsuperscript{\rm 1}Shanghai Artificial Intelligence Laboratory,
    \textsuperscript{\rm 2}Shanghai Jiao Tong University
}

\begin{document}

\maketitle

\begin{abstract}
Existing benchmarks for large language models (LLMs) predominantely focus on assessing their capabilities through verifiable tasks. Such objective and static benchmarks offer limited utility for practical LLM selection, making it difficult for users to find suitable models for their individual needs. To bridge this gap, we present the first \textbf{User-Centric Subjective Leaderboard (USL)}, which provides a preference-driven, dynamic ranking of LLMs across diverse real-world scenarios.
Our work is built upon a thorough investigation of real human preference data, involving more than 10K subjective queries. Our investigation reveals significant diversity and contradictions in human preferences, which limit the effectiveness of state-of-the-art reward models. To address this, we introduce \textbf{Customizable Reward Models (CRMs)}. With only 4B parameters, our CRM surpasses the performance of leading models such as GPT-4.1 and Gemini-2.5-pro, showing exceptional generalization capabilities across new topics and criteria. The USL, powered by CRMs, exhibits strong negative correlations to contradictory preferences.
\end{abstract}

\begin{links}
    \link{Project}{https://github.com/JiaQiSJTU/UserCentricLeaderboard}
\end{links}

\section{Introduction}
\label{sec:intro}

Leaderboards are crucial for establishing convincing LLMs rankings and have showcased continuous breakthroughs in recent years. They gauge a range of capabilities, including knowledge~\cite{rein2024gpqa,hendrycksmeasuring}, mathematics~\cite{hendrycks2measuring,glazer2024frontiermath}, coding~\cite{jainlivecodebench,wang2025ojbench}, safety~\cite{liang2023holistic,ren2025mask}, and etc. However, most focus on verifiable tasks, overlooking creative scenarios that involve subjective preferences~\cite{wang-etal-2024-user}. While arena-based evaluations~\cite{chiang2024chatbot} and LLM-as-a-judge benchmarks~\cite{li2024crowdsourced,alpaca_eval} have gained traction as they complement objective leaderboards by collecting human preferences from online users or employing LLMs as annotators to rate model responses, they unfortunately only reflect the aggregated preferences of the general public.
A critical gap remains: none of them cater to the individual need of real users in their daily lives, a need that is becoming increasingly urgent as AI technologies grow more pervasive.

\begin{figure}[ht]
    \centering
    \includegraphics[width=0.95\linewidth]{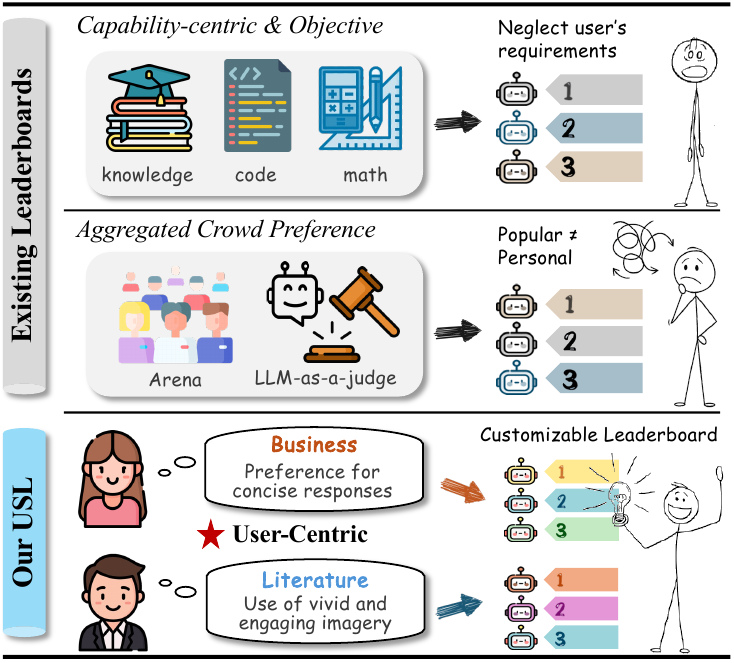}
    \caption{Comparison of USL with existing leaderboards.}
    \label{fig:intro}
\end{figure}

Building a user-centric subjective leaderboard faces two major challenges.
First, gathering model rankings from users inherently demands that each individual compare responses from dozens of models across hundreds of personal prompts to establish a stable ranking. This process is not only privacy-sensitive, but also practically impossible due to prohibitive annotation costs.
Second, unlike objective questions that are easily evaluated based on correctness, assessing responses to subjective or creative queries is inherently tough. Previous approaches have relied on reward models or directly queried LLMs for ratings. Yet, all of them demonstrate lower accuracy on the PPE preference benchmark~\cite{frick2024rlhf}, which features more subjective instances, compared to other correctness-oriented benchmarks~\cite{lambert2025rewardbench,liu2025rm}. 
Furthermore, reward models exhibit no scaling benefits~\cite{wang2025worldpm} in either data or model size on such tasks.

Considering human-LLM interactions composed of prompts and responses, we disentangle ``user-centric'' into two dimensions: (1) the variation in user interests across topics reflected by prompts, and (2) preferences for responses generated by different LLMs. We cluster subjective prompts through a human-in-the-loop pipeline, categorizing them into 12 major topics and 87 fine-grained topics. By randomly sampling from each fine-grained class, we have built \textbf{DailyBench}, a benchmark of 522 queries. This benchmark covers a broad spectrum of subjective queries, enabling users to pick topics aligned with their personal interests. Within this framework, we eliminate the need for labor-intensive manual ranking of LLMs, making it possible to leverage pairwise datasets~\cite{chiang2024chatbot,liu2025skywork} to model human preferences across different responses.

An in-depth analysis of human preferences was conducted, leveraging over 10K preference annotations from subjective and creative topics collected by the LMArena platform~\cite{chiang2024chatbot}. 
A re-evaluation by annotators reveals that despite 78.8\% of the data having a designated winner, a substantial 92\% of instances contain responses where both options are considered acceptable from a third-party viewpoint. 
Building on this, potential preference criteria are automatically extracted using LLMs. Our analysis shows no distinct distribution of criteria exists between chosen and rejected responses, and identified conflicting criteria, demonstrating the lack of unified human preferences for subjective tasks. This insight explains the suboptimal performance of reward models and LLM judges on our test sets. 

Rather than implicitly modeling human preferences, we highlight the importance of explicitly conditioning on preference criteria for subjective scenarios. To address this, we propose Customizable Reward Models (CRMs). Our experiments indicate that the LLM-as-a-judge approach falls short in effectively utilizing specified criteria and shows significant internal bias. In contrast, our CRMs, trained on smaller LLMs, deliver superior performance, achieving 97.27\% accuracy on preference recognition tasks compared to GPT-4.1's 91.96\% accuracy. Furthermore, recognizing potential mismatches between criteria and responses pairs when applying CRMs to our User-centric Subjective Leaderboard (USL), we introduce three distinct noising strategies to the given criteria. The results show that CRMs maintain robust performance across diverse evaluation scenarios.

By integrating CRMs, our USL gives users unprecedented control. They can not only filter prompts by topic to focus on areas of interest, but also define personalized preference criteria for evaluation. Following Arena Hard~\cite{li2024crowdsourced}, we chose gemini-2.0-flash-001 as the baseline and compute win rates against it for LLMs. On the one hand, the reliability of USL is bolstered by the high accuracy of CRMs on preference recognition tasks. 
On the other hand, our examination of LLM rankings under varied criteria revealed strong negative correlations when models were evaluated against contradictory criteria, for example, Kendall's $\tau = -0.83$ ($p<0.001$) for length preferences. These findings confirm that USL successfully adapts to diverse user preferences instead of converging on a single "optimal" ranking.

To sum up, our contributions are as follows:
\begin{itemize}
    \item We introduce the first User-centric Subjective Leaderboard (USL), enabling dynamic LLM rankings customizable to individual user preferences and needs.
    
    \item Through an in-depth analysis of human preferences on subjective queries, we develop novel Customizable Reward Models (CRMs) via automatic preference mining, with model sizes ranging from 0.6B to 8B parameters.

    \item Extensive experiments demonstrate that our CRMs outperform leading models in preference recognition and lead to adaptable and reliable LLM rankings for USL.
    
\end{itemize}

\section{Related Work}
\label{sec:related_work}
We contextualize our work with existing approaches to LLM leaderboards and reward modeling.

\subsection{LLM Leaderboards}

Prior LLM benchmarking generally falls into two categories. 
The first category assesses diverse capabilities of LLMs. For example, \citet{rein2024gpqa} introduced GPQA containing expert-level questions designed to evaluate scientific knowledge.
OJBench~\cite{wang2025ojbench} was proposed to rigorously evaluate reasoning skills through competition-level programming problems. These works concentrate on advancing the boundaries of LLM abilities through verifiable and objective tasks, lacking consideration of the practical usage experience in everyday scenarios.
Another line of research complements this perspective by integrating human evaluation into the assessment process. Arena-based evaluations collect human preferences through online platforms and rank LLMs using the Elo rating systems~\cite{chiang2024chatbot}. These leaderboards encompass a number of applications such as chat, web development, and web search
, and are regarded as the gold standard for LLM rankings~\cite{ni2024mixeval}. To alleviate annotation burden and facilitate reproducible evaluations, benchmarks employing LLMs as judges to emulate human annotators have been widely adopted, such as Arena-Hard~\cite{li2024crowdsourced}, AlpacaEval~\cite{alpaca_eval} and WildBench~\cite{lin2025wildbench}. Nevertheless, these works treat humans as a homogeneous group, reporting only aggregated crow preferences.

Conversely, we tackle subjective queries prevalent in daily life and deliver customizable leaderboards for each user.

\subsection{Reward Models}

Reward models serve as indispensable proxies for providing human preference signals throughout the LLM reinforcement learning pipeline. Most existing work focuses on training reward models grounded in widely accepted principles, such as HHH~\cite{askell2021general}, or strives to learn a unified human preference via extensive data collection~\cite{xu2025magpie}. \citet{wang2024helpsteer} highlights the importance of annotator agreement for filtering high-quality preference data, while \citet{liu2025skywork} leverages an iterative cleaning process to distill superior training data. Prominent reward models continue to achieve breakthroughs on various benchmarks, including RewardBench~\cite{lambert2025rewardbench}, RM-Bench~\cite{liu2025rm}, JudgeBench~\cite{tan2025judgebench}, and etc. Nevertheless, these benchmarks prioritize correctness.
Advancement on subjective benchmarks~\cite{frick2024rlhf} is constrained, with recent work~\cite{wang2025worldpm} indicating an absence of scaling trends for such preferences. 

A limited body of work acknowledged the divergence in human preferences~\cite{zhang2024diverging} and incorporated richer contextual information in reward modeling~\cite{pitis2024improving}. 
In contrast to these approaches, which typically depend on synthetic data with pre-defined preference criteria~\cite{yu2025rewardanything}, our work directly derives criteria from authentic human preference data and extends the utility of reward models to underpin user-centric leaderboards.

\section{Preliminary Analysis of Subjective Preferences}
\label{sec:preliminary}

We analyze human preference data from the LMArena platform~\footnote{https://huggingface.co/datasets/lmarena-ai/arena-human-preference-100k}, specifically selecting samples categorized as ``creativity'' or ``real-world'' scenarios that do not necessitate ``technical-accuracy'. After filtering non-English samples and those labeled ``tie (both bad)'', we obtain 10,794 samples with effective model responses, denoted as dataset $D$. Each instance in $D$ is represented as:

\begin{equation}
    (q, o^{\rm A}, o^{\rm B}, y)
\end{equation}
where $q$ denotes the user's initial query. $o^{\rm A}$ and $o^{\rm B}$ represent responses from two competing models, which can be either single-turn or multi-turn conversations. $y\in \{\text{win}, \text{tie}, \text{lose}\}$ indicates the preference relationship between $o^{\rm A}$ and $o^{\rm B}$, as originally annotated by users. Our data reveals 78.8\% of samples exhibit clear preferences($y\neq \text{tie}$). The notation $\overline{y}$ signifies the reversed preference judgement.

\textbf{Are $o^{\rm A}$ and $o^{\rm B}$ both acceptable? Yes.} 
To validate our data beyond the original human annotations, we employed two independent annotators assessed the quality of responses across 50 randomly sampled instances from dataset $D$. 
From this third-party perspective, over 92\% of cases, both responses are deemed acceptable. This observation supports the intuition that subjective scenarios permit diverse valid responses, yet raises question about what criteria humans genuinely prioritize when expressing preferences.

\textbf{Automatic criteria extraction with LLMs.} We leverage the leading proprietary model, GPT-4o~\cite{hurst2024gpt}, to conduct an in-depth analysis of subjective human preferences. Our approach involves feeding GPT-4o the tuple $(q, o^{\rm A}, o^{\rm B})$. The model is instructed to objectively analyze the strengths and weaknesses of both responses, and subsequently hypothesize potential preference criteria from two angles, i.e., $o^{\rm A}\succ o^{\rm B}$ and $o^{\rm A}\prec o^{\rm B}$. The extracted criteria must be articulated as high-level statements, free from sample-specific details, yielding $c^{\rm A}$ and $c^{\rm B}$ respectively.

\textbf{Does there exist distribution discrepancy between criteria for chosen and rejected responses? No.}
We categorize the extracted criteria into two sets: $s_{\rm chosen}$ comprising criteria for responses identified as winners by $y$, and $s_{\rm rejected}$, derived from those marked by $\overline{y}$. Using Qwen-3-Embedding-8B~\cite{qwen3embedding}, we collect embeddings for each criterion and subsequently apply K-means clustering to automatically group all criteria into two clusters. The resulting Adjusted Rand Index score of 0.001 reveals no distinct separation between these sets. In other words, criteria favored in some instances may be disfavored in others, a finding corroborated by manual inspection. For example, while some users prefer detailed and lengthy responses, others favor concise and direct answers.

In summary, our analysis confirms \textbf{the absence of unified human preferences for subjective queries}, which motivates our development of customizable reward models in subsequent sections. Analyses of criteria characteristics and reward models' performances are in Sec.~\ref{sec:setup} and Sec.~\ref{sec:results}.

\section{User-centric Subjective Leaderboard}
\label{sec:approach}

\begin{figure*}
    \centering
    \includegraphics[width=0.9\linewidth]{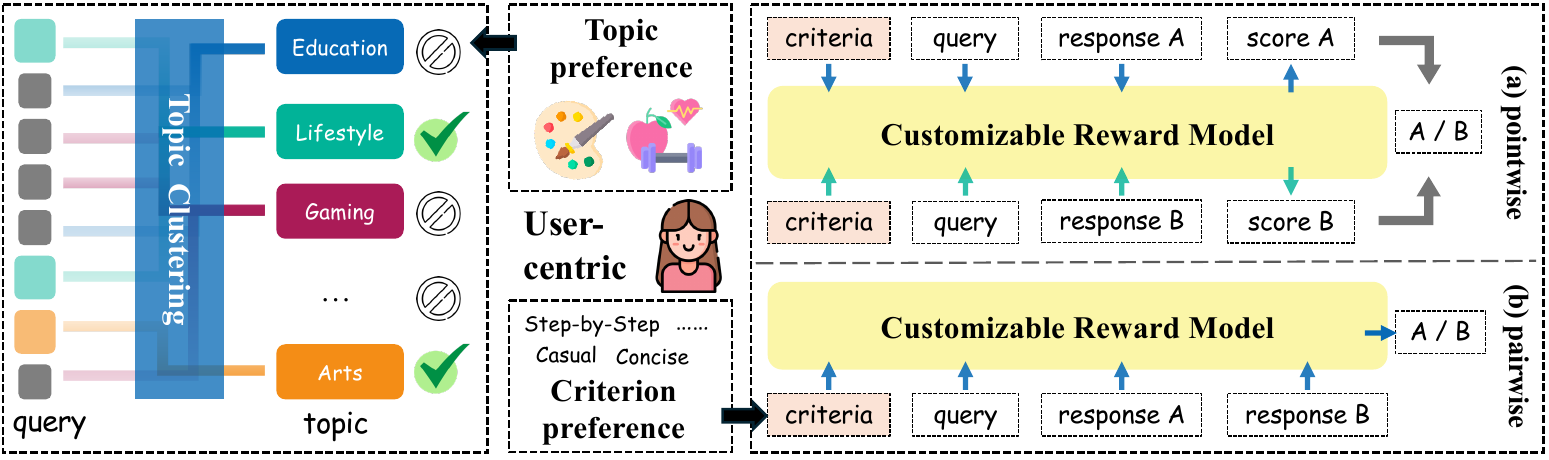}
    \caption{An illustration of the approach for USL.}
    \label{fig:approach}
\end{figure*}

The USL leverages a collection of real user queries and computes LLM win rates against a baseline model, consistent with Arena-Hard~\cite{li2024crowdsourced}. Unlike static benchmarks that reflect aggregated crowd preferences, the USL generates dynamic rankings by incorporating two disentangled dimensions of user preference as illustrated in Fig~\ref{fig:approach}: (1) \textbf{topic preference:} Reflected by the query. We employ the clutering pipeline described in Sec.\ref{sec:topic_clustering} to ensure comprehensive coverage of subjective topics, allowing users to focus on areas of interest. (2) \textbf{criteria preference:} Pertaining to response evaluation. We introduce customizable reward model in Sec.\ref{sec:customizable_rm}, enabling USL to integrate user-specified evaluation criteria and dynamically recompute LLM win rates. Interface screenshots are available in the Appendix. 

\subsection{Topic Clustering}
\label{sec:topic_clustering}

Topic clustering aims to both enhance comprehension of subjective tasks and enable users to focus on specific scenarios within the USL. We organize user queries from $D$ into a hierarchical classification tree using a hybrid approach that combines automated algorithms with human supervision. Specifically, we implement a five-stage clustering pipeline, building upon BERTopic~\cite{grootendorst2022bertopic} and following Arena Explorer~\cite{tang2025explorer}: 
\begin{itemize}
    \item Encode each query in $D$ into a dense 4096-dimensional representation using the Qwen-3 embedding model~\cite{qwen3embedding}.
    
    \item Compress the dimensionality into 5 by UMAP~\cite{mcinnes2018umap} based on a cosine similarity metric.
    \item Cluster the queries by HDBSCAN~\cite{mcinnes2017hdbscan} into groups with a minimum of 20 queries.
    \item Reassign outliers with both a conservative strategy ``c-TF-IDF'' and a comprehensive strategy ``distributions''.
    \item Extract the representative queries in each cluster to summarize the topics by prompting LLMs.
\end{itemize}
The pipeline can be iteratively applied to derive hierarchical categories. We adopted a human-in-the-loop approach, incorporating human supervision after each iteration. This process enabled us to identify 87 fine-grained topics, which were then manually categorized into 12 classes, ranging from Society \& Politics to Art \& Culture.

To build a subjective benchmark that accurately reflects diverse real-world scenarios while ensuring computational feasibility for online evaluation, we randomly selected 6 user queries from each fine-grained topic cluster. Consequently, our benchmark, \textbf{DailyBench}, comprises 522 authentic user queries, comparable in scale to Arena-Hard-Auto~\cite{li2024crowdsourced}. Responses from various LLMs are pre-collected for these queries, which will be assessed using our CRMs.

In the current implementation, USL enables users to select evaluation cases by choosing topic clusters. In the future, we will also consider additional test cases together with a similarity-based activation mechanism for more precise awareness on topic preferences.

\subsection{Customizable Reward Model}
\label{sec:customizable_rm}

\subsubsection{Problem Formulation}
Customizable reward modeling extends conventional reward modeling by incorporating explicit preference criteria $c$ for output evaluation.
Given extracted criteria, each sample in $D_c$ is derived from $D$ as: 
\begin{equation}
    (c, q, o^{\rm A}, o^{\rm B}, y_c)
\end{equation}
where the binary label $y_c\in \{0, 1\}$ is criterion-dependent:
\begin{align}
y_c = \begin{cases}
0 & \text{if } c = c^{\rm A} \\
1 & \text{if } c = c^{\rm B}.
\end{cases}
\end{align}

Equivalently, this formulation can be expressed as:
\begin{equation}
    (c, q, o^{\rm chosen}, o^{\rm reject})
\end{equation}
with:
\begin{align}
(o^{\rm chosen}, o^{\rm rejected}) = \begin{cases}
(o^{\rm A}, o^{\rm B}) & \text{if } c = c^{\rm A} \\
(o^{\rm B}, o^{\rm A}) & \text{if } c = c^{\rm B}.
\end{cases}
\end{align}

\subsubsection{Training Objective}
To acquire criteria-specific preferences, we fine-tuned LLM-based reward models $r_{\theta}$ employing two different training objectives.

Following~\cite{ouyang2022training}, we utilize the conventional Bradley-Terry model with a pairwise ranking loss:
\begin{equation}
    \mathcal{L}_{\rm ranking} = -\log (\sigma(r_{\theta}(c, q, o^{\rm chosen}) - r_{\theta}(c, q, o^{\rm reject})))
\end{equation}

Although reward models trained with $\mathcal{L}_{\rm ranking}$ can competently rank responses based on the provided criteria and query, their \textit{pointwise} evaluation treats each response independently. This can lead to an oversight of subtle comparative features present between response pairs. To mitigate this issue, we re-conceptualized the task as a binary classification problem, incorporating cross-entropy loss.

\begin{equation}
\begin{aligned}
    \hat{y}_c &= \sigma(r_{\theta}(c, q, o^{\rm A}, o^{\rm B})), \\
    \mathcal{L}_{\rm cls} &= - y_c \log \hat{y}_c - (1-y_c)\log(1- \hat{y}_c).  
\end{aligned}
\label{eq:cls}
\end{equation}
$r_{\theta}$ executes a \textit{pairwise} comparison, assessing both candidate responses concurrently.

\subsubsection{Noising Strategies}

While the criteria and chosen responses are well-aligned in $D_c$, user-provided criteria in USL may not always effectively identify the preferred response in arbitrary pairs. We cluster the extracted criteria with the BERTopic algorithm, similar to the method describled in Sec.~\ref{sec:topic_clustering}. Then, we define $\overline{c}$ as criteria leading to a flipped $y_c$, and introduce three nosing strategies:
\begin{itemize}
    \item \textbf{Criteria Removal:} We randomly drop criteria from $c$ to simulate sparse user preference statements. 
    \item \textbf{Criteria Addition:} We incorporate extraneous criteria, i.e., criteria outside the clusters covered in $\overline{c}$, to test robustness against unhelpful and redundant preferences. 
    \item \textbf{Criteria Replacement:} We substitutes criteria with the ones from $\overline{c}$, modeling cases where both responses contain desireble attributes. 
\end{itemize}
It should be noted that each criterion is considered equally important in shaping the response. We preserve the majority of original criteria in $c$ to prevent label flipping, ensuring that $y_c$ remains unchanged.
These noising strategies serve two key purposes: (1) constructing more challenging test sets for rigorous evaluation of CRMs' performance in USL, and (2) augmenting training data to enhance model robustness against imperfect or ambiguous user criteria.

\section{Experiment Setup}
\label{sec:setup}

\subsection{Dataset}

As outlined in Sec.~\ref{sec:preliminary}, we collect 10,794 real human preference data. The label distribution shows 38.39\% for $o^A \succ o^B$, 40.39\% for $o^B \succ o^A$, and 21.22\% ties, indicating a near-even distribution of winning labels between two options.

\begin{figure}
    \centering
    \subfigure[Topic Distribution]{
    \centering
    \includegraphics[width=0.98\linewidth]{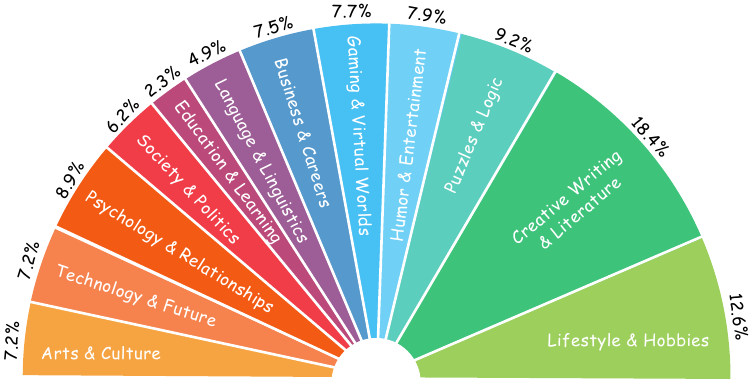}
    }
    
    \subfigure[Criterion Distribution]{
    \centering
    \includegraphics[width=0.97\linewidth]{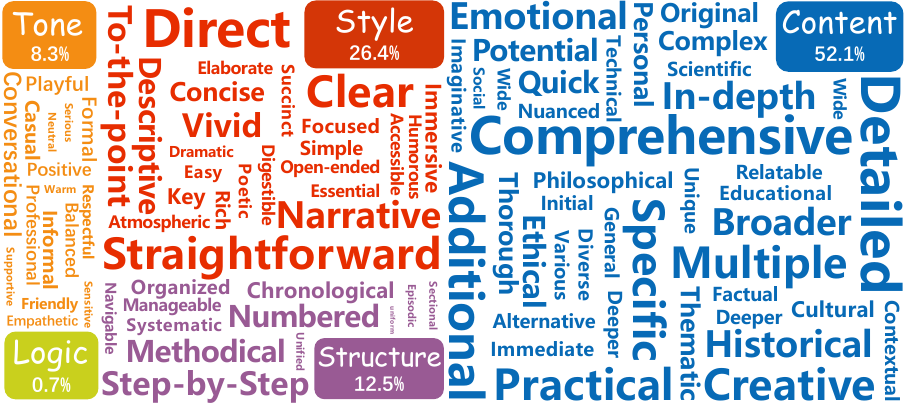}
    \label{fig:distribution-criterion}
    }
    \caption{The distribution of query topics and criteria. High-frequency adjectives are selected to characterize distinct criteria clusters.}
    \label{fig:distribution}
\end{figure}

\textbf{Data Categorization:} Building on our hierarchical topic classification from Sec.~\ref{sec:topic_clustering}, we analyze the distribution of queries cross 12 major categories (excluding 40 outliers), as visualized in Fig.~\ref{fig:distribution}. The most prevalent category is ``Creative Writing \& Literature'' with 1,979 samples, accounting for 18\% of online subjective user queries. We also iteratively cluster the extracted criteria into detailed and broad classes. Subsequently, we prompt GPT-4o to classify each detailed class into the 5 categories defined by~\citet{yu2025rewardanything}. As Fig.~\ref{fig:distribution-criterion} illustrates, our extracted criteria primarily focus on content (52.14\%), followed by style (26.35\%) and structure (12.54\%). Our approach not only captures the natural distribution of preference principle for response comparison, but also yields 95,511 unique criteria, substantially more diverse than the pre-defined 250 principles in prior work.

\textbf{Train-Test Split:} Following previous work~\cite{lambert2025rewardbench,liu2025rm}, we validate the effectiveness of CRMs in the preference recognition task, i.e., selecting the preferred response from two candidates.
We split $D_c$ into training and test sets at an approximate 9:1 ratio, leveraging the cluster labels to construct specialized test subsets for topic generalization ${\rm T}$ and criterion generalization ${\rm C}$. 980 samples from $8$ fine-grained topics and 793 samples from $3$ broad criterion categories are reserved for test. 

By incorporating criteria $c$, each instance in $D$ is converted into two samples with opposing preference labels. We construct complementary subsets $D^{+}$ and $D^{-}$ containing samples aligned with the original human preference $y$ and its reverse $\overline{y}$, respectively. To evaluate CRMs' robustness in practical scenarios, we apply noising strategies to the more challenging $D^{-}$ subset. See more statistics in the Appendix.

\subsection{Baselines \& Implementation Details}

We evaluated reward models (RMs) spanning a wide performance spectrum from RewardBenchV2~\cite{malik2025rewardbench}, including 4 RMs from Skywork~\cite{liu2025skywork} and 3 from AllenAI~\cite{lambert2024tulu3,malik2025rewardbench}.
Additionally, we assessed state-of-the-art LLMs as judges under two settings: with and without criteria conditioning. The models include GPT-4.1, Gemini-2.5~\cite{comanici2025gemini} and various sized models from Qwen3~\cite{yang2025qwen3}.

We implement CRMs using Qwen3 backbone models ranging from 0.6B to 8B parameters. For training, 10\% of the training data is randomly held out for validation. All models are trained with a batch size of 32 for a maximum of 3 epochs, using a learning rate of 1e-5 with cosine annealing scheduling. Models are evaluated on the validation set every 50 steps and early stopping with a patience of 3 is applied to prevent overfitting. We preserve the model parameters corresponding to the checkpoint achieving the lowest validation loss. All experiments were conducted on a single A800 GPU with 80GB of memory. We'll release all data and code licensed under CC-BY-NC-4.0.
\begin{table}[t]
    \centering
    \scriptsize
    \begin{tabular}{l|c|ccc}
    \toprule[1pt]
    \textbf{Models}  &  RB-V2 & $D^{\rm T+}$ & $D^{\rm C+}$ & Avg \\
    \midrule[1pt]
    \rowcolor{light-gray0} Skywork-Reward-V2-Llama-3.1-8B & \textbf{84.1} & \textbf{67.9} & \textbf{69.4} & \textbf{68.7} \\
    Skywork-Reward-V2-Qwen3-8B     & 78.4 & 66.7 & 67.0 & 66.9 \\
    \rowcolor{light-gray0} Skywork-Reward-V2-Llama-3.2-3B & 74.7 & 66.0 & 67.8 & 66.9 \\
    Skywork-Reward-Llama-3.1-8B-v0.2 & 71.8 & 60.7 & 63.4 & 62.1 \\
    \rowcolor{light-gray0} Llama-3.1-Tulu-3-8B-SFT-RM-RB2 & 68.2 & 59.4 & 60.5 & 60.0 \\
    Llama-3.1-8B-Base-RM-RB2 & 64.9 & 60.8 & 61.2 & 61.0 \\
    \rowcolor{light-gray0} Llama-3.1-Tulu-3-8B-RM & 59.0 & 60.2 & 64.3 & 62.3 \\
    \bottomrule[1pt]
    \end{tabular}
     \caption{Accuracy (\%) of specialized RMs on RewardBench V2 (RB-V2) and our test sets. ``Avg'' calculates the mean accuracy on $D^{\rm T+}$ and $D^{\rm C+}$.}
    \label{tab:rm-comparison}
\end{table}

\begin{figure}[t]
    \centering
    \includegraphics[width=0.98\linewidth]{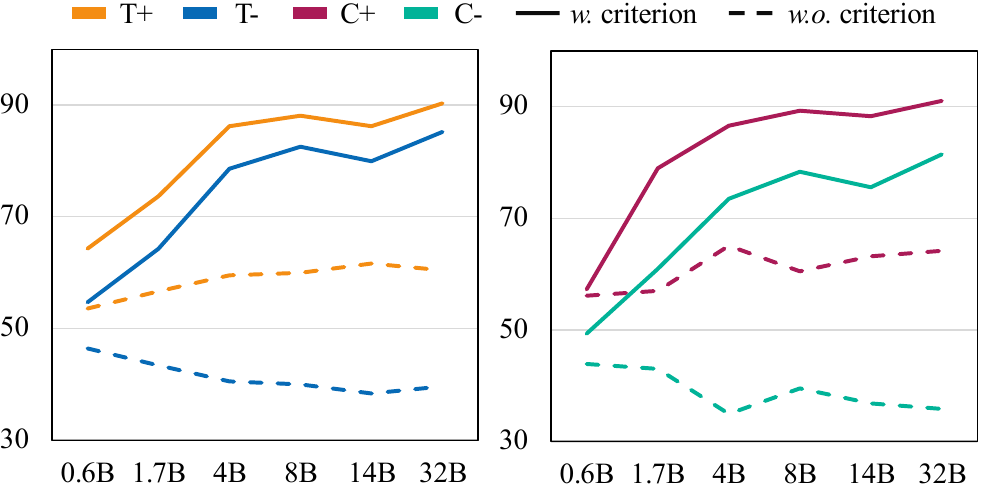}
    \caption{Accuracy(\%) for using 0.6B to 32B Qwen3 models as a judge with or without conditioning on criteria.}
    \label{fig:qwen-criterion}
\end{figure}

\section{Results and Analysis}
\label{sec:results}

\begin{table*}[h!]
    \centering
    \small
    \begin{tabular}{l|ccccc|c|ccccc|c}
    \toprule[1pt]
    & \multicolumn{6}{c|}{\textit{\textbf{topic generalization}}} & \multicolumn{6}{c}{\textit{\textbf{criterion generalization}}} \\
    \textbf{Models}    & $D^{\rm T+}$ & $D^{\rm T-}$ & $D^{\rm T-}_{\rm remove}$ & $D^{\rm T-}_{\rm add}$ & $D^{\rm T-}_{\rm replace}$ & Avg & $D^{\rm C+}$ & $D^{\rm C-}$ & $D^{\rm C-}_{\rm remove}$ & $D^{\rm C-}_{\rm add}$ & $D^{\rm C-}_{\rm replace}$ & Avg \\
    \midrule[1pt]
    \rowcolor{light-gray0} GPT-4.1 & 95.7 & 89.3 & 83.5 & 82.2 & 72.1 & 84.6 & 94.7 & 88.2 & 81.6 & 80.7 & 71.5 & 83.3 \\
    Gemini-2.5-Pro & 95.1 & 83.3 & 74.8 & 75.1 & 63.8 & 78.4 & 93.1 & 82.5 & 71.8 & 72.8 & 63.9 & 76.8 \\
    \rowcolor{light-gray0} Qwen3-32B & 90.3 & 85.2 & 80.5 & 80.1 & 74.3 & 82.1 & 91.1 & 81.5 & 75.9 & 75.0 & 70.1 & 78.7\\
    Qwen3-14B & 86.2 & 79.9 & 73.0 & 73.2 & 67.1 & 75.9 & 88.3 & 75.5 & 71.1 & 68.4 & 61.8 & 73.0 \\
    \rowcolor{light-gray0} Qwen3-8B & 88.1 & 82.6 & 75.2 & 77.2 & 71.3 & 78.9 & 89.3 & 78.3 & 69.7 & 69.2 & 64.2 & 74.1 \\
    Qwen3-4B & 86.2 & 78.6 & 72.8 & 73.0 & 67.9 & 75.7 & 86.6 & 73.5 & 66.2 & 64.6 & 57.8 & 69.7 \\
    \rowcolor{light-gray0} Qwen3-1.7B & 73.7 & 64.3 & 59.1 & 60.2 & 56.8 & 62.8 & 78.9 & 60.9 & 54.4 & 60.8 & 51.0 & 61.2\\
    Qwen3-0.6B & 64.3 & 54.7 & 52.9 & 50.4 & 51.3 & 54.7 & 57.3 & 49.3 & 44.1 & 47.4 & 46.9 & 49.0 \\
    \midrule
    \rowcolor{light-gray0} CRM-8B & \textbf{97.9} & 97.2 & 93.8 & 95.3 & 92.2 & 95.3 & 97.2 & 96.3 & \textbf{93.7} & 94.0 & \textbf{90.7} & 94.4 \\
    CRM-4B & 97.0 & \textbf{97.5} & \textbf{94.1} & \textbf{95.9} & \textbf{93.5} & \textbf{95.6} & \textbf{97.6} & \textbf{97.0} & 93.3 & \textbf{94.6} & 90.5 &  \textbf{94.6} \\
    \rowcolor{light-gray0} CRM-1.7B & 92.9 & 92.9 & 87.7 & 89.9 & 87.7 & 90.2 & 93.3 & 92.3 & 88.3 & 88.9 & 83.9 & 89.3 \\
    CRM-0.6B & 68.4 & 63.8 & 57.8 & 60.1 & 57.7 & 61.6 & 56.6 & 53.5 & 48.7 & 49.6 & 50.5 & 51.8 \\
    \bottomrule[1pt]
    \end{tabular}
    \caption{Accuracy (\%) of LLM judges and CRMs on different test subsets. ``Avg'' refers to the averaged performance on 5 test sets for topic generalization and criterion generalization correspondingly.}
    \label{tab:crm_main_result}
\end{table*}

\subsection{RM Performance without Criteria}

We selected 7 reward models with varying performances on RewardBenchV2, ranging from 59.0\% to 84.1\%. 
Table\ref{tab:rm-comparison} compares their performance on RewardBenchV2's correctness-oriented tasks against our newly built test sets, which focus on subjective daily scenarios. 
The evaluated reward models show comparable performance on DailyBench, from 60.0 to 68.7\%, despite significant differences on RewardBenchV2. 
Although Llama-3.1-Tulu-3-8B-RM trails Skywork-Reward-Llama-3.1-8B-v0.2 by 12.8\% on RewardBenchV2, they exhibit nearly identical performance on DailyBench. Notably, even the top-performing Skywork-Reward-V2-Llama-3.1-8B cannot surpass the 70\% accuracy ceiling on subjective tasks.
This aligns with the finding that human preference labeling agreement is capped at 70$\sim$80\%~\cite{wang2024helpsteer,cui2023ultrafeedback}.

The empirical evidence leads us to conclude that\textbf{ human preference patterns are effectively characterized by the Pareto principle (80/20 rule)}, which manifests as an approximate 70:30 ratio in subjective tasks. This finding demonstrate the absence of unified human preferences and establishes that understanding heterogeneous subjective preferences is critical, complementing to traditional correctness-oriented evaluation. Rather than treating preference diversity as noise to be eliminated, we content that reward modeling should actively leverage this diversity through exploitation of varied user preferences.

\subsection{LLM-as-a-Judge Performance with Criterion}

We evaluate LLMs as judges under two distinct settings: explicit criteria-conditioned preference recognition, and implicit aggregated crowd preference assessment. To analyze scaling trends, we test Qwen3 series models across varying parameter sizes. The results are depicted in Fig.~\ref{fig:qwen-criterion}.
The introduction of explicit criteria enables LLM judges to successfully handle bidirectional preference modeling ($\rm T+/T-$ and $\rm C+/C-$ test sets). Thus, LLM judges circumvent the limitations of the 80/20 preference distribution, confirming that both responses can be acceptable when evaluated against different preference criteria.

Despite explicit criterion provision, Fig.~\ref{fig:qwen-criterion} reveals persistent performance gaps: evaluation accuracy on $D^{\rm T-}$ and $D^{\rm C-}$ consistently lags behind $D^{\rm T+}$ and $D^{\rm C+}$ by significant margins. While Qwen3 models demonstrate robust instruction-following capabilities, their alignment process appears to internalize aggregated crowd preferences. This internalization creates systematic bias during judgment tasks, causing deviations from specified evaluation criteria.

\subsection{Performance of CRMs}
Table~\ref{tab:crm_main_result} presents the comparative performance of our fine-tuned CRMs against zero-shot prompted LLM judges when provided with explicit criteria.

\textbf{LLMs demonstrate strong criteria-based preference recognition capabilities.} Among proprietary models, GPT-4.1 achieves superior  accuracy over Gemini-2.5-Pro by 6.5\% and 6.2\% on respective test subsets. The open-source Qwen3 series shows progressively better performance with increasing model size. Notably, Qwen-3-32B not only outperforms Gemini-2.5-Pro but also reaches competitive accuracy with GPT-4.1, indicating there is no conspicuous gap between open-source and proprietary models. Both topic and criterion generalization follow similar performance trends, though the latter presents marginally greater challenges.

\textbf{CRMs outperform state-of-the-art LLM judges.} Through supervised fine-tuning for customizable reward modeling, CRMs achieve superior performance, reaching around 95\% accuracy and demonstrating significant improvements over zero-shot baselines. For example, CRM-4B boosts Qwen3-4B's accuracy from 75.7\% to 95.6\% in topic generalization, and from 69.7\% to 94.6\% in criterion generalization. Remarkably, our CRM with merely 4B parameters surpasses GPT-4.1 by over 11\% accuracy. 

\textbf{CRMs exhibit enhanced robustness and reduced bias.} 
The fine-tuning process effectively mitigates performance disparities between $(D^{\rm T+}, D^{\rm T-})$ and $(D^{\rm C+}, D^{\rm C-})$ test sets, with CRMs achieving remarkably balanced accuracy with gap less than 1\% compared to GPT-4.1's 6\% differential. Furthermore, CRMs with more than 1B parameters demonstrate consistent performance stability across all noising strategies described in Sec.~\ref{sec:customizable_rm}, showcasing superior adaptability to challenging application scenarios.

In summary, our CRMs show consistently superior accuracy and robustness across evaluation scenarios, establishing their effectiveness for LLM ranking in the USL.

\begin{table*}[]
    \centering
    \small
    \begin{tabular}{l|ccccc|c|ccccc|c}
    \toprule[1pt]
    & \multicolumn{6}{c|}{\textit{\textbf{topic generalization}}} & \multicolumn{6}{c}{\textbf{\textit{criterion generalization}}} \\
    \textbf{Models}    & $D^{\rm T+}$ & $D^{\rm T-}$ & $D^{\rm T-}_{\rm remove}$ & $D^{\rm T-}_{\rm add}$ & $D^{\rm T-}_{\rm replace}$  & Avg & $D^{\rm C+}$ & $D^{\rm C-}$ & $D^{\rm C-}_{\rm remove}$ & $D^{\rm C-}_{\rm add}$ & $D^{\rm C-}_{\rm replace}$ & Avg \\
    \midrule[1pt]
    \rowcolor{light-gray0} zero-shot    & 59.5 & 40.5 & - & - & - & 50.0 & 65.1 & 34.9 & - & - & - & 50.0 \\
    fine-tuning w. $\mathcal{L}_{\rm cls}$ & 64.6 & 35.4 & - & - & - & 50.0 & 62.3 & 37.7 & - & - & - & 50.0 \\
    \rowcolor{light-gray0} zero-shot w. $c$ & 86.2 & 78.6 & 72.8 & 73.0 & 67.9 & 75.7 & 86.6 & 73.5 & 66.2 & 64.6 & 57.8 & 69.7 \\
    \midrule[1pt]
    CRM  & 97.0 & \textbf{97.5} & \textbf{94.1} & \textbf{95.9} & \textbf{93.5} & \textbf{95.6} & \textbf{97.6} & \textbf{97.0} & 93.3 & \textbf{94.6} & \textbf{90.5} & \textbf{94.6} \\
    \rowcolor{light-gray0} \quad w.o. noises & \textbf{97.5} & 96.2 & 92.4 & 93.8 & 89.5 & 93.9 & 96.9 & 97.0 & \textbf{94.0} & 93.1 & 87.8 & 93.8 \\
    \quad w. $\mathcal{L}_{\rm ranking}$ & 78.4 & 76.0 & 67.2 & 68.3 & 65.6 & 71.1 & 75.4 & 75.2 & 70.1 & 68.2 & 65.8 & 70.9 \\

    \bottomrule[1pt]
    \end{tabular}
    \caption{Ablation study conducted on CRM-4B. }
    \label{tab:crm_ablation}
\end{table*}

\subsection{Ablation Study for CRMs}

We analyze CRM-4B's technical design through ablations in Table~\ref{tab:crm_ablation}. The baseline approach, "fine-tuning w. $\mathcal{L}_{\rm cls}$", which trains Qwen3-4B via Eq.~\ref{eq:cls} without criteria conditioning, shows limited improvement on $D^{\rm T+}$ and degraded performance on $D^{\rm C+}$ compared to zero-shot prompting. This result confirms the absence of superficial preference biases in our dataset, as the model fails to learn meaningful patterns.

Our experiments identify random noise injection combined with $\mathcal{L}_{\rm cls}$ optimization as the most effective training strategy. Ablation studies reveal that noise-free training yields inferior results, particularly reducing robustness to criterion replacement scenarios. While $\mathcal{L}_{\rm ranking}$ demonstrates stronger criterion generalization than zero-shot approaches, it underperforms on topic generalization tasks and significantly lags behind  $\mathcal{L}_{\rm cls}$. The latter benefits from joint response evaluation in a single forward pass, an architectural advantage that enables token-level comparison between candidates under given criteria, rather than relying solely on scalar outputs.
Based on analysis of the extracted criteria in Fig.~\ref{fig:distribution}, we recognize the inherent challenges in assigning absolute scores to subjective responses. For instance, preferences like ``detailed responses with in-depth analysis'' fundamentally require relative assessment rather than absolute quantification, as they lack universal evaluation standards.

\subsection{Case Study of USL}

We collect responses for DailyBench from 12 leading LLMs, besides using Gemini-2.0-Flash-001 as baseline. Using reward models, we compare each LLM's performance against the baseline and compute their corresponding win rate(\%). The default ranking $\varnothing$ is determined by Skywork-Reward-V2-Llama-3.1-8B. As shown in Table~\ref{tab:case_study}, DeepSeek-R1 ranks highest, whereas Claude-3.7-Sonnet performs weakest according to aggregated crowd preferences.

We re-rank models for USL according to four different preference criteria as follows:
\begin{enumerate}
    \item \textit{Prefer in-depth exploration and detailed analysis.}
    \item \textit{Preference for concise responses that are easy to read. }
    \item \textit{Deliver a creative and inspiring narrative tone.}
    \item \textit{Provide a step-by-step structure.}
\end{enumerate}
LLMs ranked by CRM-4B are shown in Table~\ref{tab:case_study}. While previous research has treated stylistic attributes (e.g., response length) as confounding factors that may compromise ranking accuracy, we conceptualize these as legitimate dimensions of subjective preference that appeal to different humans. 
$c_1$ favors DeepSeek-R1 that generates more elaborate responses, and $c_2$ prefers Claude-3.7-Sonnet for conciseness. Gemini-2.5-Pro adopts a more engaging tone, while DeepSeek-R1 exhibits a propensity for structured formats.

We calculate the correlations among the rankings produced by $\varnothing$, $c_1$ and $c_2$. The Kendall correlation between $\varnothing$ and $c_1$ is $0.43$ ($p<0.05$), suggesting that conventional reward models exhibit a systematic bias toward longer responses. The correlation between $c_1$ and $c_2$ is $-0.83$ with $p<-0.001$, indicating a strong negative correlation due to their nearly reversed preferences. Our USL effectively captures and operationalizes distinct user-specific preferences, generating meaningfully differentiated model rankings.

\begin{table}[t]
    \centering
    \small
    \begin{tabular}{l|ccccc}
    \toprule[1pt]
       \textbf{Model}  & $\varnothing$ & $c_1$ & $c_2$ & $c_3$ & $c_4$\\
    \midrule[1pt]
     \rowcolor{light-gray0}  DeepSeek-R1 & 1 & 1 & 12 & 3 & 1 \\
    Gemma-3-27b-it & 2 & 4 & 11 & 4 & 6 \\
 \rowcolor{light-gray0} Gemini-2.5-Flash & 3 & 6 & 7 & 7 & 4 \\
         Qwen3-32B & 4 & 7 & 6 & 6 & 3 \\
 \rowcolor{light-gray0}  Gemini-2.5-Pro  & 5 &  3 & 8 & 1 & 2 \\
     o3-2025-04-16 & 6 & 8 & 5 & 11 & 5 \\
 \rowcolor{light-gray0}          GPT-4.1 & 7 & 11 & 4 & 8 & 10 \\
o4-mini-2025-04-16 & 8 & 9 & 2 & 12 & 9 \\
 \rowcolor{light-gray0}  Qwen3-235B-A22B & 9 & 4 & 9 & 2 & 8 \\
           QwQ-32B & 10 & 2 & 10 & 5 & 7 \\
  \rowcolor{light-gray0}   o1-2024-12-17 & 11 & 9 & 3 & 9 & 12 \\
 Claude-3.7-Sonnet & 12 & 12 & 1 & 10 & 11 \\
 
    \bottomrule[1pt]
    \end{tabular}
    \caption{Rankings of LLMs. $\varnothing$ represents the averaged human preference measure by Skywork-Reward-V2-Llama-3.1-8B. $c_1$ to $c_4$ refer to rankings under 4 different criteria describing human preference with CRM-4B as the judge.}
    \label{tab:case_study}
\end{table}

\section{Conclusion}
\label{sec:conclusion}

In this work, we propose USL to provide personalized LLM rankings for subjective tasks. We disentangle user-centric LLM evaluations into two dimensions, i.e., topic preference and criteria preference, and demonstrate the importance of explicit criteria for modeling human preferences in subjective daily scenarios. Our CRMs achieve superior performance on preference recognition tasks while requiring fewer parameters. The CRM-powered USL generates reliable rankings, as evidenced by strong negative correlations with reversed preferences.

Our work represents the first step toward user-centric LLM evaluation. The current implementation allows users to select preferred topics and specify personalized criteria for dynamic LLM ranking. Valuable future directions include extending coverage to objective tasks (e.g., mathematics and coding), developing more capable CRMs for diverse scenarios, and collecting human-centric data to iteratively improve USL accuracy. Incorporating CRMs for LLM alignment also presents a promising research avenue.

\bibliography{aaai2026}

\begin{thebibliography}{38}
\providecommand{\natexlab}[1]{#1}

\bibitem[{Askell et~al.(2021)Askell, Bai, Chen, Drain, Ganguli, Henighan,
  Jones, Joseph, Mann, DasSarma et~al.}]{askell2021general}
Askell, A.; Bai, Y.; Chen, A.; Drain, D.; Ganguli, D.; Henighan, T.; Jones, A.;
  Joseph, N.; Mann, B.; DasSarma, N.; et~al. 2021.
\newblock A general language assistant as a laboratory for alignment.
\newblock \emph{arXiv preprint arXiv:2112.00861}.

\bibitem[{Chiang et~al.(2024)Chiang, Zheng, Sheng, Angelopoulos, Li, Li, Zhu,
  Zhang, Jordan, Gonzalez et~al.}]{chiang2024chatbot}
Chiang, W.-L.; Zheng, L.; Sheng, Y.; Angelopoulos, A.~N.; Li, T.; Li, D.; Zhu,
  B.; Zhang, H.; Jordan, M.; Gonzalez, J.~E.; et~al. 2024.
\newblock Chatbot arena: An open platform for evaluating llms by human
  preference.
\newblock In \emph{Forty-first International Conference on Machine Learning}.

\bibitem[{Comanici et~al.(2025)Comanici, Bieber, Schaekermann, Pasupat,
  Sachdeva, Dhillon, Blistein, Ram, Zhang, Rosen et~al.}]{comanici2025gemini}
Comanici, G.; Bieber, E.; Schaekermann, M.; Pasupat, I.; Sachdeva, N.; Dhillon,
  I.; Blistein, M.; Ram, O.; Zhang, D.; Rosen, E.; et~al. 2025.
\newblock Gemini 2.5: Pushing the frontier with advanced reasoning,
  multimodality, long context, and next generation agentic capabilities.
\newblock \emph{arXiv preprint arXiv:2507.06261}.

\bibitem[{Cui et~al.(2023)Cui, Yuan, Ding, Yao, Zhu, Ni, Xie, Liu, and
  Sun}]{cui2023ultrafeedback}
Cui, G.; Yuan, L.; Ding, N.; Yao, G.; Zhu, W.; Ni, Y.; Xie, G.; Liu, Z.; and
  Sun, M. 2023.
\newblock UltraFeedback: Boosting Language Models with High-quality Feedback.
\newblock \emph{CoRR}.

\bibitem[{Frick et~al.(2024)Frick, Li, Chen, Chiang, Angelopoulos, Jiao, Zhu,
  Gonzalez, and Stoica}]{frick2024rlhf}
Frick, E.; Li, T.; Chen, C.; Chiang, W.-L.; Angelopoulos, A.~N.; Jiao, J.; Zhu,
  B.; Gonzalez, J.~E.; and Stoica, I. 2024.
\newblock How to Evaluate Reward Models for RLHF.
\newblock arXiv:2410.14872.

\bibitem[{Glazer et~al.(2024)Glazer, Erdil, Besiroglu, Chicharro, Chen,
  Gunning, Olsson, Denain, Ho, Santos et~al.}]{glazer2024frontiermath}
Glazer, E.; Erdil, E.; Besiroglu, T.; Chicharro, D.; Chen, E.; Gunning, A.;
  Olsson, C.~F.; Denain, J.-S.; Ho, A.; Santos, E. d.~O.; et~al. 2024.
\newblock Frontiermath: A benchmark for evaluating advanced mathematical
  reasoning in ai.
\newblock \emph{arXiv preprint arXiv:2411.04872}.

\bibitem[{Grootendorst(2022)}]{grootendorst2022bertopic}
Grootendorst, M. 2022.
\newblock BERTopic: Neural topic modeling with a class-based TF-IDF procedure.
\newblock \emph{arXiv preprint arXiv:2203.05794}.

\bibitem[{Hendrycks et~al.(2021{\natexlab{a}})Hendrycks, Burns, Basart, Zou,
  Mazeika, Song, and Steinhardt}]{hendrycksmeasuring}
Hendrycks, D.; Burns, C.; Basart, S.; Zou, A.; Mazeika, M.; Song, D.; and
  Steinhardt, J. 2021{\natexlab{a}}.
\newblock Measuring Massive Multitask Language Understanding.
\newblock In \emph{International Conference on Learning Representations}.

\bibitem[{Hendrycks et~al.(2021{\natexlab{b}})Hendrycks, Burns, Kadavath,
  Arora, Basart, Tang, Song, and Steinhardt}]{hendrycks2measuring}
Hendrycks, D.; Burns, C.; Kadavath, S.; Arora, A.; Basart, S.; Tang, E.; Song,
  D.; and Steinhardt, J. 2021{\natexlab{b}}.
\newblock Measuring Mathematical Problem Solving With the MATH Dataset.
\newblock In \emph{Thirty-fifth Conference on Neural Information Processing
  Systems Datasets and Benchmarks Track (Round 2)}.

\bibitem[{Hurst et~al.(2024)Hurst, Lerer, Goucher, Perelman, Ramesh, Clark,
  Ostrow, Welihinda, Hayes, Radford et~al.}]{hurst2024gpt}
Hurst, A.; Lerer, A.; Goucher, A.~P.; Perelman, A.; Ramesh, A.; Clark, A.;
  Ostrow, A.; Welihinda, A.; Hayes, A.; Radford, A.; et~al. 2024.
\newblock Gpt-4o system card.
\newblock \emph{arXiv preprint arXiv:2410.21276}.

\bibitem[{Jain et~al.(2025)Jain, Han, Gu, Li, Yan, Zhang, Wang, Solar-Lezama,
  Sen, and Stoica}]{jainlivecodebench}
Jain, N.; Han, K.; Gu, A.; Li, W.-D.; Yan, F.; Zhang, T.; Wang, S.;
  Solar-Lezama, A.; Sen, K.; and Stoica, I. 2025.
\newblock LiveCodeBench: Holistic and Contamination Free Evaluation of Large
  Language Models for Code.
\newblock In \emph{The Thirteenth International Conference on Learning
  Representations}.

\bibitem[{Lambert et~al.(2024)Lambert, Morrison, Pyatkin, Huang, Ivison,
  Brahman, Miranda, Liu, Dziri, Lyu, Gu, Malik, Graf, Hwang, Yang, Bras,
  Tafjord, Wilhelm, Soldaini, Smith, Wang, Dasigi, and
  Hajishirzi}]{lambert2024tulu3}
Lambert, N.; Morrison, J.; Pyatkin, V.; Huang, S.; Ivison, H.; Brahman, F.;
  Miranda, L. J.~V.; Liu, A.; Dziri, N.; Lyu, S.; Gu, Y.; Malik, S.; Graf, V.;
  Hwang, J.~D.; Yang, J.; Bras, R.~L.; Tafjord, O.; Wilhelm, C.; Soldaini, L.;
  Smith, N.~A.; Wang, Y.; Dasigi, P.; and Hajishirzi, H. 2024.
\newblock Tülu 3: Pushing Frontiers in Open Language Model Post-Training.

\bibitem[{Lambert et~al.(2025)Lambert, Pyatkin, Morrison, Miranda, Lin, Chandu,
  Dziri, Kumar, Zick, Choi et~al.}]{lambert2025rewardbench}
Lambert, N.; Pyatkin, V.; Morrison, J.; Miranda, L. J.~V.; Lin, B.~Y.; Chandu,
  K.; Dziri, N.; Kumar, S.; Zick, T.; Choi, Y.; et~al. 2025.
\newblock RewardBench: Evaluating Reward Models for Language Modeling.
\newblock In \emph{Findings of the Association for Computational Linguistics:
  NAACL 2025}, 1755--1797.

\bibitem[{Li et~al.(2024)Li, Chiang, Frick, Dunlap, Wu, Zhu, Gonzalez, and
  Stoica}]{li2024crowdsourced}
Li, T.; Chiang, W.-L.; Frick, E.; Dunlap, L.; Wu, T.; Zhu, B.; Gonzalez, J.~E.;
  and Stoica, I. 2024.
\newblock From Crowdsourced Data to High-Quality Benchmarks: Arena-Hard and
  BenchBuilder Pipeline.
\newblock \emph{arXiv preprint arXiv:2406.11939}.

\bibitem[{Li et~al.(2023)Li, Zhang, Dubois, Taori, Gulrajani, Guestrin, Liang,
  and Hashimoto}]{alpaca_eval}
Li, X.; Zhang, T.; Dubois, Y.; Taori, R.; Gulrajani, I.; Guestrin, C.; Liang,
  P.; and Hashimoto, T.~B. 2023.
\newblock AlpacaEval: An Automatic Evaluator of Instruction-following Models.

\bibitem[{Liang et~al.(2023)Liang, Bommasani, Lee, Tsipras, Soylu, Yasunaga,
  Zhang, Narayanan, Wu, Kumar, Newman, Yuan, Yan, Zhang, Cosgrove, Manning, Re,
  Acosta-Navas, Hudson, Zelikman, Durmus, Ladhak, Rong, Ren, Yao, WANG,
  Santhanam, Orr, Zheng, Yuksekgonul, Suzgun, Kim, Guha, Chatterji, Khattab,
  Henderson, Huang, Chi, Xie, Santurkar, Ganguli, Hashimoto, Icard, Zhang,
  Chaudhary, Wang, Li, Mai, Zhang, and Koreeda}]{liang2023holistic}
Liang, P.; Bommasani, R.; Lee, T.; Tsipras, D.; Soylu, D.; Yasunaga, M.; Zhang,
  Y.; Narayanan, D.; Wu, Y.; Kumar, A.; Newman, B.; Yuan, B.; Yan, B.; Zhang,
  C.; Cosgrove, C.~A.; Manning, C.~D.; Re, C.; Acosta-Navas, D.; Hudson, D.~A.;
  Zelikman, E.; Durmus, E.; Ladhak, F.; Rong, F.; Ren, H.; Yao, H.; WANG, J.;
  Santhanam, K.; Orr, L.; Zheng, L.; Yuksekgonul, M.; Suzgun, M.; Kim, N.;
  Guha, N.; Chatterji, N.~S.; Khattab, O.; Henderson, P.; Huang, Q.; Chi,
  R.~A.; Xie, S.~M.; Santurkar, S.; Ganguli, S.; Hashimoto, T.; Icard, T.;
  Zhang, T.; Chaudhary, V.; Wang, W.; Li, X.; Mai, Y.; Zhang, Y.; and Koreeda,
  Y. 2023.
\newblock Holistic Evaluation of Language Models.
\newblock \emph{Transactions on Machine Learning Research}.

\bibitem[{Lin et~al.(2025)Lin, Deng, Chandu, Ravichander, Pyatkin, Dziri,
  Le~Bras, and Choi}]{lin2025wildbench}
Lin, B.~Y.; Deng, Y.; Chandu, K.; Ravichander, A.; Pyatkin, V.; Dziri, N.;
  Le~Bras, R.; and Choi, Y. 2025.
\newblock WildBench: Benchmarking LLMs with Challenging Tasks from Real Users
  in the Wild.
\newblock In \emph{The Thirteenth International Conference on Learning
  Representations}.

\bibitem[{Liu et~al.(2025{\natexlab{a}})Liu, Zeng, Xiao, He, Liu, Wang, Yan,
  Shen, Zhang, Xu et~al.}]{liu2025skywork}
Liu, C.~Y.; Zeng, L.; Xiao, Y.; He, J.; Liu, J.; Wang, C.; Yan, R.; Shen, W.;
  Zhang, F.; Xu, J.; et~al. 2025{\natexlab{a}}.
\newblock Skywork-Reward-V2: Scaling Preference Data Curation via Human-AI
  Synergy.
\newblock \emph{arXiv preprint arXiv:2507.01352}.

\bibitem[{Liu et~al.(2025{\natexlab{b}})Liu, Yao, Min, Cao, Hou, and
  Li}]{liu2025rm}
Liu, Y.; Yao, Z.; Min, R.; Cao, Y.; Hou, L.; and Li, J. 2025{\natexlab{b}}.
\newblock RM-Bench: Benchmarking Reward Models of Language Models with Subtlety
  and Style.
\newblock In \emph{The Thirteenth International Conference on Learning
  Representations}.

\bibitem[{Malik et~al.(2025)Malik, Pyatkin, Land, Morrison, Smith, Hajishirzi,
  and Lambert}]{malik2025rewardbench}
Malik, S.; Pyatkin, V.; Land, S.; Morrison, J.; Smith, N.~A.; Hajishirzi, H.;
  and Lambert, N. 2025.
\newblock RewardBench 2: Advancing Reward Model Evaluation.
\newblock \emph{arXiv preprint arXiv:2506.01937}.

\bibitem[{McInnes, Healy, and Astels(2017)}]{mcinnes2017hdbscan}
McInnes, L.; Healy, J.; and Astels, S. 2017.
\newblock hdbscan: Hierarchical density based clustering.
\newblock \emph{Journal of Open Source Software}, 2(11): 205.

\bibitem[{McInnes et~al.(2018)McInnes, Healy, Saul, and
  Gro{\ss}berger}]{mcinnes2018umap}
McInnes, L.; Healy, J.; Saul, N.; and Gro{\ss}berger, L. 2018.
\newblock UMAP: Uniform Manifold Approximation and Projection.
\newblock \emph{Journal of Open Source Software}, 3(29).

\bibitem[{Ni et~al.(2024)Ni, Xue, Yue, Deng, Shah, Jain, Neubig, and
  You}]{ni2024mixeval}
Ni, J.; Xue, F.; Yue, X.; Deng, Y.; Shah, M.; Jain, K.; Neubig, G.; and You, Y.
  2024.
\newblock Mixeval: Deriving wisdom of the crowd from llm benchmark mixtures.
\newblock \emph{Advances in Neural Information Processing Systems}, 37:
  98180--98212.

\bibitem[{Ouyang et~al.(2022)Ouyang, Wu, Jiang, Almeida, Wainwright, Mishkin,
  Zhang, Agarwal, Slama, Ray et~al.}]{ouyang2022training}
Ouyang, L.; Wu, J.; Jiang, X.; Almeida, D.; Wainwright, C.; Mishkin, P.; Zhang,
  C.; Agarwal, S.; Slama, K.; Ray, A.; et~al. 2022.
\newblock Training language models to follow instructions with human feedback.
\newblock \emph{Advances in neural information processing systems}, 35:
  27730--27744.

\bibitem[{Pitis et~al.(2024)Pitis, Xiao, Le~Roux, and
  Sordoni}]{pitis2024improving}
Pitis, S.; Xiao, Z.; Le~Roux, N.; and Sordoni, A. 2024.
\newblock Improving context-aware preference modeling for language models.
\newblock \emph{Advances in Neural Information Processing Systems}, 37:
  70793--70827.

\bibitem[{Rein et~al.(2024)Rein, Hou, Stickland, Petty, Pang, Dirani, Michael,
  and Bowman}]{rein2024gpqa}
Rein, D.; Hou, B.~L.; Stickland, A.~C.; Petty, J.; Pang, R.~Y.; Dirani, J.;
  Michael, J.; and Bowman, S.~R. 2024.
\newblock Gpqa: A graduate-level google-proof q\&a benchmark.
\newblock In \emph{First Conference on Language Modeling}.

\bibitem[{Ren et~al.(2025)Ren, Agarwal, Mazeika, Menghini, Vacareanu, Kenstler,
  Yang, Barrass, Gatti, Yin et~al.}]{ren2025mask}
Ren, R.; Agarwal, A.; Mazeika, M.; Menghini, C.; Vacareanu, R.; Kenstler, B.;
  Yang, M.; Barrass, I.; Gatti, A.; Yin, X.; et~al. 2025.
\newblock The mask benchmark: Disentangling honesty from accuracy in ai
  systems.
\newblock \emph{arXiv preprint arXiv:2503.03750}.

\bibitem[{Tan et~al.(2025)Tan, Zhuang, Montgomery, Tang, Cuadron, Wang, Popa,
  and Stoica}]{tan2025judgebench}
Tan, S.; Zhuang, S.; Montgomery, K.; Tang, W.~Y.; Cuadron, A.; Wang, C.; Popa,
  R.; and Stoica, I. 2025.
\newblock JudgeBench: A Benchmark for Evaluating LLM-Based Judges.
\newblock In \emph{The Thirteenth International Conference on Learning
  Representations}.

\bibitem[{Tang, Chiang, and Angelopoulos(2025)}]{tang2025explorer}
Tang, K.; Chiang, W.-L.; and Angelopoulos, A.~N. 2025.
\newblock Arena Explorer: A Topic Modeling Pipeline for LLM Evals \& Analytics.

\bibitem[{Wang et~al.(2025{\natexlab{a}})Wang, Lin, Lu, Yu, Zhang, Huang,
  Zheng, Dang, Fan, Ren et~al.}]{wang2025worldpm}
Wang, B.; Lin, R.; Lu, K.; Yu, L.; Zhang, Z.; Huang, F.; Zheng, C.; Dang, K.;
  Fan, Y.; Ren, X.; et~al. 2025{\natexlab{a}}.
\newblock WorldPM: Scaling Human Preference Modeling.
\newblock \emph{arXiv preprint arXiv:2505.10527}.

\bibitem[{Wang et~al.(2024{\natexlab{a}})Wang, Mo, Ma, Sun, Zhang, and
  Nie}]{wang-etal-2024-user}
Wang, J.; Mo, F.; Ma, W.; Sun, P.; Zhang, M.; and Nie, J.-Y.
  2024{\natexlab{a}}.
\newblock A User-Centric Multi-Intent Benchmark for Evaluating Large Language
  Models.
\newblock In \emph{Proceedings of the 2024 Conference on Empirical Methods in
  Natural Language Processing}, 3588--3612. Association for Computational
  Linguistics.

\bibitem[{Wang et~al.(2024{\natexlab{b}})Wang, Dong, Delalleau, Zeng, Shen,
  Egert, Zhang, Sreedhar, and Kuchaiev}]{wang2024helpsteer}
Wang, Z.; Dong, Y.; Delalleau, O.; Zeng, J.; Shen, G.; Egert, D.; Zhang, J.;
  Sreedhar, M.~N.; and Kuchaiev, O. 2024{\natexlab{b}}.
\newblock Helpsteer 2: Open-source dataset for training top-performing reward
  models.
\newblock \emph{Advances in Neural Information Processing Systems}, 37:
  1474--1501.

\bibitem[{Wang et~al.(2025{\natexlab{b}})Wang, Liu, Wang, He, Gao, Diao, Chen,
  Fu, Sung, Yang et~al.}]{wang2025ojbench}
Wang, Z.; Liu, Y.; Wang, Y.; He, W.; Gao, B.; Diao, M.; Chen, Y.; Fu, K.; Sung,
  F.; Yang, Z.; et~al. 2025{\natexlab{b}}.
\newblock OJBench: A Competition Level Code Benchmark For Large Language
  Models.
\newblock \emph{arXiv preprint arXiv:2506.16395}.

\bibitem[{Xu et~al.(2025)Xu, Jiang, Niu, Deng, Poovendran, Choi, and
  Lin}]{xu2025magpie}
Xu, Z.; Jiang, F.; Niu, L.; Deng, Y.; Poovendran, R.; Choi, Y.; and Lin, B.~Y.
  2025.
\newblock Magpie: Alignment Data Synthesis from Scratch by Prompting Aligned
  LLMs with Nothing.
\newblock In \emph{The Thirteenth International Conference on Learning
  Representations}.

\bibitem[{Yang et~al.(2025)Yang, Li, Yang, Zhang, Hui, Zheng, Yu, Gao, Huang,
  Lv et~al.}]{yang2025qwen3}
Yang, A.; Li, A.; Yang, B.; Zhang, B.; Hui, B.; Zheng, B.; Yu, B.; Gao, C.;
  Huang, C.; Lv, C.; et~al. 2025.
\newblock Qwen3 technical report.
\newblock \emph{arXiv preprint arXiv:2505.09388}.

\bibitem[{Yu et~al.(2025)Yu, Zeng, Gu, Wang, Wang, Meng, Zhou, Zhang, Zhang,
  and Ye}]{yu2025rewardanything}
Yu, Z.; Zeng, J.; Gu, W.; Wang, Y.; Wang, J.; Meng, F.; Zhou, J.; Zhang, Y.;
  Zhang, S.; and Ye, W. 2025.
\newblock RewardAnything: Generalizable Principle-Following Reward Models.
\newblock \emph{arXiv preprint arXiv:2506.03637}.

\bibitem[{Zhang et~al.(2024)Zhang, Wang, Hwang, Dong, Delalleau, Choi, Choi,
  Ren, and Pyatkin}]{zhang2024diverging}
Zhang, M.~J.; Wang, Z.; Hwang, J.~D.; Dong, Y.; Delalleau, O.; Choi, Y.; Choi,
  E.; Ren, X.; and Pyatkin, V. 2024.
\newblock Diverging Preferences: When do Annotators Disagree and do Models
  Know?
\newblock \emph{arXiv preprint arXiv:2410.14632}.

\bibitem[{Zhang et~al.(2025)Zhang, Li, Long, Zhang, Lin, Yang, Xie, Yang, Liu,
  Lin, Huang, and Zhou}]{qwen3embedding}
Zhang, Y.; Li, M.; Long, D.; Zhang, X.; Lin, H.; Yang, B.; Xie, P.; Yang, A.;
  Liu, D.; Lin, J.; Huang, F.; and Zhou, J. 2025.
\newblock Qwen3 Embedding: Advancing Text Embedding and Reranking Through
  Foundation Models.
\newblock \emph{arXiv preprint arXiv:2506.05176}.

\end{thebibliography}

\newpage
\appendix

\section{Ethical Consideration}

Our research is based on a publicly available dataset sourced from the LMArena platform. We acknowledge that the original dataset may contain content that could be sensitive to certain groups. However, since addressing such content falls outside the scope of this study, we did not perform specialized filtering of these samples. For our analysis of user preferences, we employed GPT-4o, ensuring an impartial comparison without introducing additional sensitive information. We will release our data licensed under CC-BY-NC-4.0, which permits only non-commercial use and is intended exclusively for research purposes.

\section{The screenshots of USL}

\begin{figure*}
    \centering
    \includegraphics[width=0.8\linewidth]{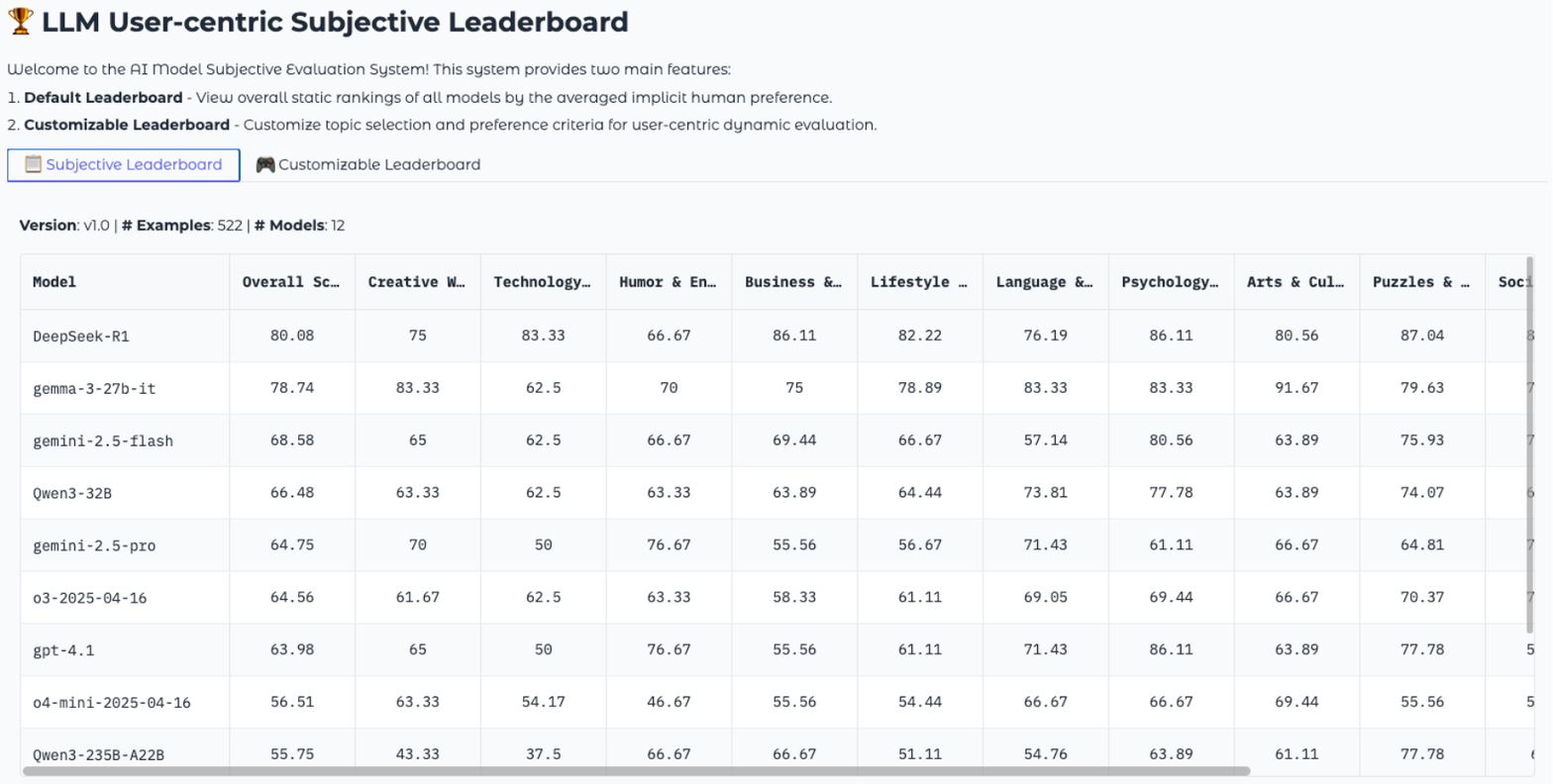}
    \caption{The screenshot for the static ranking of LLMs on all queries in USL.}
    \label{fig:static_screenshot}
\end{figure*}

\begin{figure*}
    \centering
    \includegraphics[width=0.85\linewidth]{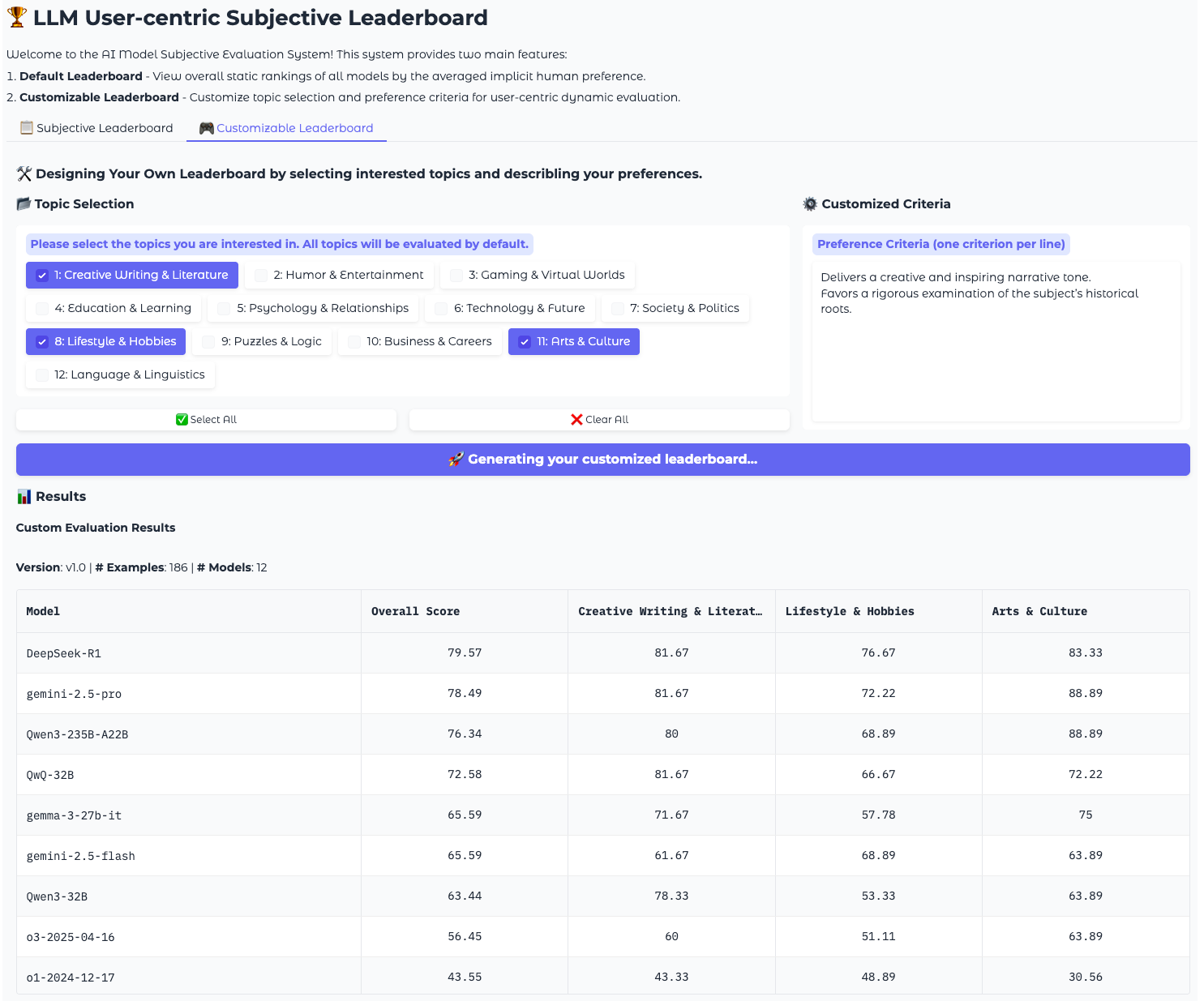}
    \caption{The screenshot for the customizable ranking of LLMs on selected topics and provided preference criteria for USL.}
    \label{fig:customized_screenshot}
\end{figure*}

We present the screenshots of our USL. Fig.~\ref{fig:static_screenshot} displays the default static LLM ranking, computed using Skywork-Reward-V2-Llama-3.1-8B. Fig.~\ref{fig:customized_screenshot} illustrates a personalized LLM ranking tailored to a user with interests in three topics: Creative Writing \& Literature, Lifestyle \& Hobbies, and Arts \& Culture. The user prefers the responses that deliver a creative and inspiring narrative tone, and provide a rigorous examination of the subject's historical roots. Based on these user-centric information, the USL recommends DeepSeek-R1, Gemini-2.5-pro and Qwen3-235B-A22B as the top 3 LLMs for this user.

\section{Data Statistics}

Table~\ref{tab:statistics} summarizes the statistics of our test sets, including the number of samples, the average number of criteria per sample, the average number of turns and the label distribution. Our test sets contain multi-turn conversations, with an average of approximately 2.5 turns per dialogue (including the initial user query). The average number of criteria per sample varies from 1.9 to 6.7 across different subsets. Additionally, we compute the distribution for different labels, which indicate a relatively balanced ratio. Therefore, to reduce computational overhead, we do not perform order swapping between response candidates, and instead rely on the original randomized order for evaluation.

\begin{table}[h]
    \centering
    \small
    \begin{tabular}{l|cccc}
    \toprule[1pt]
    \textbf{Subset} & \textbf{\#Samples} & \textbf{\#Turns} & \textbf{\#Criteria} & \textbf{Label Ratio} \\
    \midrule[1pt]
    \rowcolor{light-gray0} $D^{\rm T+}$ & 980 & 2.7 & 4.5 & 466:514\\
    $D^{\rm T-}$ & 980 & 2.7 & 4.5 & 514:466 \\
    \rowcolor{light-gray0} $D^{\rm T+}_{\rm remove}$ & 980 & 2.7 & 1.9 & 514:466 \\
    $D^{\rm T+}_{\rm add}$ & 980 & 2.7 & 6.4 & 514:466 \\
    \rowcolor{light-gray0} $D^{\rm T+}_{\rm replace}$ & 980 & 2.7 & 4.5 & 514:466\\
    \midrule[1pt]
    $D^{\rm C+}$ & 793 & 2.5 & 4.7 & 440:353 \\
    \rowcolor{light-gray0} $D^{\rm C-}$ & 793 & 2.5 & 4.7 & 353:440 \\
    $D^{\rm C+}_{\rm remove}$ & 793 & 2.5 & 2.0 & 353:440 \\
    \rowcolor{light-gray0} $D^{\rm C+}_{\rm add}$ & 793 & 2.5 & 6.7 & 353:440 \\
    $D^{\rm C+}_{\rm replace}$ & 793 & 2.5 & 4.7 & 353:440 \\
    \bottomrule[1pt]
    \end{tabular}
    \caption{Statistics for test sets.}
    \label{tab:statistics}
\end{table}

\section{Analysis for Criteria}

To analyze the most representative criteria within each broad cluster, we extracted adjectives and ranked them by frequency. When an adjective appeared in multiple clusters, we retained it only in the cluster with the highest frequency. Table~\ref{tab:criteria} presents the top 30 adjectives for each cluster. Our results demonstrate that the extracted criteria encompass not only a wide range of aspects but also diverse linguistic expressions. Additionally, they reflect contrasting human preferences, such as specific vs. broad, concise vs. elaborate, and formal vs. informal.

Logic contains fewer criteria, accounting for only 0.7\% of the total. This observation can be attributed to that our study focuses on subjective scenarios, where users prioritize inspiration and diverse perspectives over objective correctness, diminishing the role of logical rigor. The analysis by GPT-4.1 indicates that responses to subjective topics exhibit minimal variation in logical structure compared to other dimensions, further reducing its prominence in our extracted criteria.

In this work, we focus exclusively on positive descriptions of preference criteria. Future work will incorporate negative criteria (i.e., user dislikes) to develop a more comprehensive and robust CRM. Additionally, we plan to enhance criteria quality through refined filtering operations and more granular comparative analyses.

\begin{table}[]
    \small
    \centering
    \begin{tabular}{c|p{5cm}}
    \toprule[1pt]
        \textbf{Category} &  \textbf{Top-30 Adjectives} \\
    \midrule[1pt]
         \rowcolor{light-gray0} Content & detailed, comprehensive, additional, specific, practical, multiple, creative, emotional, historical, in-depth, broader, potential, thematic, personal, cultural, ethical, quick, complex, thorough, actionable, philosophical, imaginative, educational, diverse, immediate, original, different, unique, deeper, broad \\
        \hline
        Style & direct, clear, straightforward, narrative, vivid, unnecessary, to-the-point, descriptive, concise, key, immersive, focused, rich, simple, poetic, succinct, easy, essential, excessive, humorous, metaphorical, digestible, elaborate, minimalistic, stylistic, atmospheric, accessible, evocative, open-ended, extra \\
        \hline
         \rowcolor{light-gray0} Structure & structured, logical, step-by-step, organized, well-structured, methodical, numbered, systematic, well-organized, lyrical, distinct, easy-to-follow, problem-solving, chronological, decision-making, grammatical, coherent, linear, structural, manageable, bullet-point, segmented, rhythmic, repetitive, sequential, decisive, categorical, persuasive, labeled, list-based \\
        \hline
        Tone & conversational, formal, playful, positive, professional, balanced, friendly, empathetic, neutral, consistent, respectful, supportive, light-hearted, motivational, sensitive, casual, informal, enthusiastic, reflective, whimsical, approachable, empathy, controversial, understanding, serious, warm, optimistic, entertaining, negative, lighthearted\\
        \hline
         \rowcolor{light-gray0} Logic & mathematical, unsupported, geometric, non-existent, question-and-answer, unwarranted, well-constructed, error-free, user-corrected, tied \\
    \bottomrule[1pt]
    \end{tabular}
    \caption{Frequent adjectives in extracted criteria under each category.}
    \label{tab:criteria}
\end{table}

\section{Performances for Case Study}

The specific win rate (\%) of each LLMs in the case study is shown in Table~\ref{tab:case_study_score}.

\begin{table}[h]
    \centering
    \small
    \begin{tabular}{l|ccccc}
    \toprule[1pt]
       \textbf{Model}  & $\varnothing$ & $c_1$ & $c_2$ & $c_3$ & $c_4$\\
    \midrule[1pt]
 \rowcolor{light-gray0}  DeepSeek-r1 & 80.1 & 80.3 & 25.5 & 73.4 & 71.5 \\
                         Gemma-3-27b-it & 78.7 & 69.2 & 27.0 & 68.0 & 58.4 \\
 \rowcolor{light-gray0}  Gemini-2.5-flash & 68.6 & 65.9 & 37.4 & 64.8 & 62.6 \\
                         Qwen3-32B & 66.5 & 63.6 & 42.2 & 65.9 & 64.2 \\
 \rowcolor{light-gray0}  Gemini-2.5-pro & 64.8 & 70.1 & 35.4  & 75.7 & 65.3  \\
                         o3-2025-04-16 & 64.6 & 51.3 & 57.1 & 46.2 & 58.6 \\
 \rowcolor{light-gray0}  GPT-4.1 & 64.0 & 34.9 & 63.8 & 55.8 & 51.0 \\
                         o4-mini-2025-04-16 & 56.5 & 37.8 & 69.7 & 43.1 & 51.5 \\
 \rowcolor{light-gray0}  Qwen3-235B-A22B & 55.8 & 69.2 & 33.5 & 74.3 & 53.1 \\
                         QwQ-32B & 53.3 & 72.6 & 31.4 & 67.2 & 55.4 \\
  \rowcolor{light-gray0} o1-2024-12-17 & 52.7 & 36.8 & 66.3 & 48.3 & 38.3 \\
                         Claude-3.7-sonnet & 46.9 & 33.7 & 70.7 & 46.9 & 42.3 \\
    \bottomrule[1pt]
    \end{tabular}
    \caption{Win rate (\%) of LLMs evaluated by reward models. $\varnothing$ represents the averaged human preference measure by Skywork-Reward-V2-Llama-3.1-8B. $c_1$ to $c_4$ refer to rankings under 4 different criteria describing explicit human preference with CRM-4B as the judge.}
    \label{tab:case_study_score}
\end{table}

\end{document}